\documentclass{article}

\usepackage{amssymb}
\usepackage{amsmath}
\newtheorem{theorem}{Theorem}
\newtheorem{lemma}{Lemma}

\newtheorem{proof}{Proof}
\usepackage[justification=centering]{caption}
\usepackage{multirow}
\usepackage{caption} 
\usepackage{bbm}

\usepackage{times}
\usepackage{graphicx} 
\usepackage{subfigure} 

\usepackage{natbib}

\usepackage{algorithm}
\usepackage{algorithmic}

\usepackage{hyperref}

\usepackage[accepted]{icml2017}

\icmltitlerunning{Directly and Efficiently Optimizing Prediction Error and AUC of Linear Classifiers}

\begin{document} 

\twocolumn[
\icmltitle{Directly and Efficiently Optimizing Prediction Error \\ and AUC of Linear Classifiers}



\icmlsetsymbol{equal}{*}

\begin{icmlauthorlist}
\icmlauthor{Hiva Ghanbari}{equal,to}
\icmlauthor{Katya Scheinberg}{equal,to}
\end{icmlauthorlist}

\icmlaffiliation{to}{Lehigh University, Bethlehem, PA, USA}

\icmlcorrespondingauthor{Hiva Ghanbari}{hiva.ghanbari@gmail.com}
\icmlcorrespondingauthor{Katya Scheinberg}{katyascheinberg@gmail.com}

\icmlkeywords{Black-Box Optimization, DFO, Bayesian Optimization, AUC, RBF-Kernel SVM}

\vskip 0.3in
]



\printAffiliationsAndNotice{}  

\begin{abstract} 
The predictive quality of machine learning models is typically measured in terms of their (approximate) expected prediction error or the so-called Area Under the Curve (AUC) for a particular data distribution. However, when the models are constructed by the means of  empirical risk minimization, surrogate functions such as the logistic loss are optimized instead. This is done  because the empirical approximations  of the expected  error and AUC functions are nonconvex and nonsmooth, and more importantly have zero derivative almost everywhere. In this work, we show that in the case of linear predictors, and under the assumption that the data has normal distribution, the expected error and the expected AUC are not only smooth, but have closed form expressions, which depend on the first and second moments of the normal distribution. 
Hence, we derive derivatives of these two functions and use these derivatives in an optimization algorithm to directly optimize the expected error and the AUC. 
In the case of real data sets, the derivatives can be  approximated using empirical  moments. We show that even  when  data is not normally distributed, computed derivatives are sufficiently useful to render an efficient optimization method and high quality solutions. Thus, we propose a gradient-based optimization method for direct optimization of the prediction error and AUC. Moreover, the per-iteration complexity of the proposed algorithm has no dependence on the size of the data set, unlike those for optimizing logistic regression and all other well known empirical risk minimization problems.  
\end{abstract} 

\section{Introduction} \label{sec.introduction}
In this paper, we consider classical binary linear classification problems in supervised Machine Learning (ML). In other words, given  a finite set
 labeled data (labeled to form a positive and a negative class),  the aim is to obtain a linear classifier that  predicts the positive/negative labels of unseen data points as accurately as possible. 
To measure the accuracy of a classifier, the expected prediction error, which measures the percentage of mislabeled data points, also known as the $0-1$ loss function, is often used. However, since the empirical approximation of the prediction error is a nonsmooth nonconvex function, whose gradient is either zero or not defined. Hence, other surrogate loss functions are typically used to determine the linear classifier. For example, standard ML tools, such as support vector machines \cite{cortes2,Osuna,Smola} and logistic regression \cite{Hosmer}, aim to optimize empirical prediction error, while using hinge loss and logistic loss, respectively, as surrogate functions  of the $0-1$ loss function. 

Many real world ML problems are dealing with imbalanced data sets, which contain rare positive data points, as the minority class, but numerous negative ones, as the majority class. When these two data classes are highly imbalanced, the prediction error function is not a useful prediction measure. For example, if the data set contains only $0.01\%$ of the positive examples, then a predictor that simply classifies every data point as  negative has $99.99\%$ accuracy, while obviously failing to achieve any meaningful prediction.  The prediction measure is often modified to incorporate class importance weights, in which case it can be used for imbalanced data sets. All results of this paper easily extend to such modification. However, a more established and robust measure of prediction accuracy which is used in practice is Area Under Receiver Operating Characteristic (ROC) Curve (AUC) \cite{hanley}.
AUC is a reciprocal of the ranking loss, which is similar to the $0-1$ loss, in the sense that it measures the percentage of pairs of data samples, one from the negative class and one from the positive class, such that the classifier assigns a larger label to the negative sample than to the positive one. In other words, $1-$AUC counts the percentage of incorrectly ``ranked" pairs \cite{mann}. 
The empirical approximation of AUC, just as that of $0-1$ loss, is a discontinuous, nonsmooth function, whose gradient is either zero or undefined. 
 This difficulty motivates various techniques for optimizing continuous approximations of AUC.
For example, the ranking  loss can be replaced by convex loss functions such as  pairwise logistic loss or hinge loss \cite{Joachims,steck,rudin,jin}, which results in continuous convex optimization problem. A drawback of such approach, aside from the fact that a different objective is optimized,  is that such loss has to be computed for each {\em pair} of data points, which significantly increases the  complexity of the underlying optimization algorithm.  

In this paper, we propose a novel method  of directly optimizing the expected prediction error and the expected AUC value of a linear classifier in the case of binary classification problems.
First, we use the probabilistic interpretation of the expected prediction error and we show that  if the distribution of the positive and negative classes obey  normal distributions, then the expected prediction error of a linear classifier is a smooth function with a closed-form expression. Thus, its gradient can be computed and a gradient-based optimization algorithm can be used. The closed form of the function depends on  the first and second moments of the related normal distributions, hence these moments are needed to compute the function value as well as the gradient.

Similarly, under the assumption that the class of the positive and negative data sets jointly obey a normal distribution, we show that the corresponding expected AUC value of a linear classifier is a smooth function with closed form expression, which depends on the first and second moments of the distribution. Similarly to optimizing the prediction error, this novel result allows any gradient-based optimization algorithm to be applied  to optimize the AUC value of a linear classifier. 

Through empirical experiments we show that even when the data sets do not obey normal distribution, optimizing the derived functional forms of prediction error and AUC, using empirical approximate moments, often produces better predictors than those obtained by optimizing surrogate approximations, such as logistic and hinge losses. This behavior is in contrast with, for example, Linear Discriminant Analysis (LDA) \cite{LDA},  which is the method to compute linear classifiers under the Gaussian assumption. 


Another key advantage of the proposed method over the classical empirical risk minimization  is that the training data is only used once at the beginning of the algorithm to compute the approximate moments. After that  each iteration of an optimization algorithm only depends on the dimension of the classifier, while optimizing
logistic loss or pairwise hinge loss using gradient-based method depends on the data size at each iteration.

The paper is organized as follows. In the next section we state preliminaries and the problem description. In Section \ref{sec.ER.AUC} we show that the prediction error and AUC are smooth functions if the data obey normal distribution. We present computational results in Section \ref{sec.numerical}, and finally, we state our conclusions in Section \ref{sec.conclusion}.

\section{Preliminaries and Problem Description}\label{sec.ER.AUC}
We consider the classical setting of  \emph{supervised} machine learning, where we are given a finite \emph{training set} $\mathcal{S}$ of $n$ pairs,
\begin{equation*} \label{training.set}
\mathcal{S} := \{(x_i,y_i)~:~ i =1,\cdots,n\},
\end{equation*}
where $x_i \in \mathbb{R}^d$ are the \emph{input}  vectors of \emph{features} and $y_i \in\{+1,-1\}$ are the \emph{binary output} labels. It is assumed that each pair $(x_i,y_i)$ is an i.i.d.  sample of the random variable $(X,Y)$ with some unknown \emph{joint probability distribution} $P_{X,Y}(x,y)$ over the input space $\mathcal{X}$ and output space $\mathcal{Y}$.
The set $\mathcal{S}$ is known as a \emph{training set}. 
The goal is to compute a linear \emph{classifier function} $f: \mathcal{X} \to \mathcal{Y}$, so that given a random input variable $X$, $f$ can accurately predict the corresponding label $Y$. 

As discussed in \S \ref{sec.introduction}, there are two different performance measures to evaluate the quality of  $f$: the prediction error, which approximates  the \emph{expected risk},
 and the AUC. 
Expected risk of a linear classifier $f(x;w) = w^T x$ for \emph{0-1 loss function} is defined as
\begin{equation} \label{expectedRisk}
\begin{aligned}
F_{error}\left( w \right) &= \mathbb{E}_{\mathcal{X},\mathcal{Y}} \left[ \ell_{01} \left(f(X;w),Y\right)\right ] \\
& = \int_{\mathcal{X}} \int_{\mathcal{Y}} P_{X,Y}(x,y) \ell_{01} \left(f(x;w),y\right) dy dx,
\end{aligned}
\end{equation}
where
\begin{equation*} \label{step_0,1}
\ell_{01}\left(f(x;w),y\right) =  \begin{cases} +1 & \text {if} ~~ y \cdot f(x;w) <0,\\ 0 & \text {if} ~~ y \cdot f(x;w) \geq 0. \end{cases}
\end{equation*}

A finite sample approximation of \eqref{expectedRisk}, given a training set $\mathcal{S}$,  is the following empirical risk
\begin{equation} \label{func_0,1}
\hat F_{error}\left(w;\mathcal{S}\right) = \frac{1}{n}\sum_{i=1}^{n} \ell_{01}\left(f(x_i;w),y_i\right).
\end{equation}

The difficulty of optimizing  \eqref{func_0,1}, even approximately, arises from the fact that its gradient is either not defined or is equal to zero. Thus, gradient-based optimization methods cannot be applied. The most common alternative is to  utilize the \emph{logistic regression loss function}, as an approximation of the prediction error and solve the following  unconstrained convex optimization problem
\begin{equation} \label{min.logreg}
\begin{aligned}
\min_{w \in \mathbb{R}^d}  \hat F_{log}(w) =& \frac{1}{n}\sum_{i=1}^{n} \log \left(1+ \exp(-y_i \cdot f(x_i;w)) \right) \\ &+ \lambda r(w) ,
\end{aligned}
\end{equation}
where $ \lambda r(w)$ is the regularization term, with  $r(\cdot)=\|\cdot\|_1$ or $r(\cdot)=\|\cdot\|_2$ as possible examples.

We now discuss the AUC function as the quality measure of a classifier, which is often the industry standard.
For that let us define
\begin{equation*}
\begin{aligned}
\mathcal{S}^+ &:= \{x: (x,y)\in \mathcal{S},~y=+1\}\\&:=\{x_i^{+}~:~ i=1,\cdots,n^+\},~~\text{where}~~x^+_i \in \mathbb{R}^d~~\text{and} \\
\mathcal{S}^- &:= \{x: (x,y)\in \mathcal{S},~y=-1\}\\&:= \{x_j^{-}~:~ j=1,\cdots,n^-\},~~\text{where}~~x^-_j \in \mathbb{R}^d.
\end{aligned}
\end{equation*}
Hence $\mathcal{S}^+$ and $\mathcal{S}^-$ are the sets of all positive and negative samples  in  $\mathcal{S}$, respectively, 
and they contain only inputs $x$, instead of pairs $(x,y)$.   Let $|\mathcal{S}^+|=n^+$ and $|\mathcal{S}^-|=n^-$. 
The AUC value of a  classifier $f(x;w)$, given the positive set $\mathcal{S}^+$ and the negative set $\mathcal{S}^-$ can be obtained via Wilcoxon-Mann-Whitney (WMW) statistic result \cite{mann}, e.g.,
\begin{equation} \label{AUC}
\begin{aligned}
&\hat F_{AUC}\left(w;\mathcal{S}^+,\mathcal{S}^-\right) \\
&= \frac{\sum_{i=1}^{n^+} \sum_{j=1}^{n^-}  \mathbbm{1}  \left [f(x^+_i;w) > f(x^-_j;w) \right ]}{n^+  \cdot n^-},
\end{aligned}
\end{equation}
where
\begin{equation*} \label{step_AUC}
\begin{aligned}
  &\mathbbm{1} \left [ f(x_i^+;w)> f(x_j^-;w)\right ] \\ &= \begin{cases} +1 & \text {if} ~~ f(x_i^+;w)> f(x_j^-;w),\\ 0 & \text {otherwise.} \end{cases}
  \end{aligned}
\end{equation*}


Now, let $\mathcal{X}^+$ and $\mathcal{X}^-$ denote the space of the positive and negative input vectors, respectively, so that $x_i^{+}$ is an i.i.d. observation of the random variable $X^+$ from  $\mathcal{X}^+$ and $x_j^{-}$ is an i.i.d. observation of the random variable $X^-$ from $\mathcal{X}^-$. Then, given the joint probability distribution $P_{X^+,X^-}\left(x^+,x^-\right)$, the expected AUC function of a  classifier $f(x;w)$ is defined as 
\begin{equation} \label{def.AUC}
\begin{aligned}
F_{AUC}(w) =~& \mathbb{E}_{\mathcal{X}^+,\mathcal{X}^-} \left [\mathbbm{1}\left [f\left(X^+;w\right)> f\left(X^-;w\right)\right ] \right ]  \\
=~& \int_{\mathcal{X}^+} \int_{\mathcal{X}^-} P_{X^+,X^-}\left(x^+,x^-\right) \\
& \cdot \mathbbm{1}\left [f\left(x^+;w\right)> f\left(x^-;w\right)\right ] dx^- dx^+.
\end{aligned}
\end{equation}

The $\hat{F}_{AUC}\left(w;\mathcal{S}^+,\mathcal{S}^-\right)$ computed by \eqref{AUC} is an unbiased estimator of $F_{AUC}(w)$. 
Similarly to the empirical risk minimization, the problem of optimizing AUC value of a predictor is not straightforward since the gradient of this  function is either zero or not defined. Thus, gradient-based optimization methods cannot be applied. 

As in the case of prediction error, various techniques have been proposed to approximate the AUC with a surrogate function. 
In \cite{mozer}, the indicator function $ \mathbbm{1}[\cdot]$ in \eqref{AUC} is substituted with a  \emph{sigmoid surrogate function}, e.g., $1/ \left({1+e^{-\beta \left(f(x^+;w) - f(x^-;w)\right)}} \right)$ and a gradient descent algorithm is applied to this smooth approximation. The choice of the parameter $\beta$  in the  sigmoid function definition significantly affects the output of this approach; although a large value of $\beta$ renders a closer approximation of  the step function, it also results in large oscillations  of the gradients, which in turn can cause numerical issues in the gradient descent algorithm. Similarly, as is discussed in \cite{rudin}, \emph{pairwise exponential loss} and \emph{pairwise logistic loss} can be utilized as convex smooth surrogate functions of the indicator function $ \mathbbm{1}[\cdot]$. In these settings, any gradient-based optimization method can be used to optimize the resulting approximate AUC value. However, due to the required pairwise comparison of the value of $f(\cdot;w)$, for each positive and negative pair, the complexity of computing function value as well as the gradient will be of order of $\mathcal{O}\left(n^+n^- \right)$, which can be very expensive. In \cite{steck}, \emph{pairwise hinge loss} has been used as a surrogate function, resulting the following approximate AUC value

\begin{equation} \label{AUC_hinge}
\begin{aligned}
&{F}_{hinge}\left(w\right) \\ 
&= \frac{\sum_{i=1}^{n^+} \sum_{j=1}^{n^-} \max \left \{0, 1- \left( f(x^-_j;w) - f(x^+_i;w)  \right)\right \} }{n^+  \cdot n^-}.
\end{aligned}
\end{equation}
The advantage of pairwise hinge loss over other alternative approximations lies in the fact that the function values as well as the gradients of pairwise hinge loss can be computed in roughly $\mathcal{O}\left( n \log(n) \right)$ time, where $n = n^+ + n^-$, by first sorting all values $f(x^-_j;w)$ and $f(x^+_i;w)$. 
One can utilize numerous stochastic gradient schemes to reduce the per-iteration complexity of optimizing  surrogate AUC objectives, however, the approach we propose here achieves the same or better result with a simpler method. 

In this paper, we propose to optimize alternative smooth approximations of expected risk and expected AUC, which display good accuracy and also have low computational cost. Towards that end, in the next section, we show that,  if the data distribution is normal,  then the
 expected risk and expected AUC of a linear classifier are both smooth functions with closed form expressions. 

\newpage
\section{Prediction Error and AUC as Smooth Functions}\label{sec.ER.AUC} 
Consider the probabilistic interpretation of the expected error, e.g., 
\begin{equation} \label{expectedRisk_prob}
\begin{aligned}
F_{error}(w) &= \mathbb{E}_{\mathcal{X},\mathcal{Y}} \left[ \ell_{01} \left(f(X;w),Y\right)\right ] \\
& = P({Y} \cdot w^T X <0).
\end{aligned}
\end{equation}

We have the following simple lemma.

\begin{lemma} \label{ER_prob}
Given the \textit{prior probabilities} $P(Y=+1)$ and $P(Y=-1)$ we can write 
\begin{equation*}
\begin{aligned}
F_{error}(w) =~& P({Y} \cdot w^T {X} <0) \\
=~& P\left( w^T {X}^+ \leq 0\right) P\left({Y} = +1\right) \\
&+ \left(1-P\left(w^T {X}^- \leq 0\right)\right) P\left({Y} = -1\right),
\end{aligned}
\end{equation*}
where $X^+$ and $X^-$ are random variables from positive and negative classes, respectively. 
\end{lemma}

Based on the result of Lemma \ref{ER_prob},  $F_{error}(w)$ is a continuous and smooth function if   the Cumulative Distribution Function (CDF)  of the  random variable $w^T {X}$ is a continuous smooth function.  In general, it is possible to derive smoothness of the CDF of $w^T {X}$ for a variety of distributions, which will imply that in principal, continuous optimization techniques can be applied to optimize $F_{error}(w)$. However, to use gradient-based methods it is necessary to obtain an estimate of the gradient 
of $F_{error}(w)$. Here, we show that under the Gaussian assumption, gradients of $F_{error}(w)$ have a closed form expression. 

We now  state Theorem 3.3.3 from \cite{tong}, which shows that the family of multivariate Gaussian distributions is closed under linear transformations. 
\begin{theorem} \label{th.normal.close}
If $X \sim \mathcal{N}\left(\mu, \Sigma\right)$ and $X$ is in $\mathbb{R}^d$  and $Z = CX + b$, for any  $C\in \mathbb{R}^{m\times d}$  and $b\in \mathbb{R}^m$, then $Z \sim \mathcal{N}\left(C\mu+b, C\Sigma C^T\right)$.
\end{theorem}

%
In the following theorem we derive the closed form expression for the expected risk under the Gaussian assumption. 

\begin{theorem} \label{theorem.main.0_1}
Suppose that  both the positive and the negative class each obeys a normal distribution, i.e.,
\begin{equation} \label{original.normal}
{X}^+ \sim \mathcal{N}\left(\mu^+, \Sigma^{+}\right)~~~ \text{and}~~~{X}^- \sim \mathcal{N}\left(\mu^-, \Sigma^{-}\right).
\end{equation}

Then, 
\begin{equation} \label{smooth.zero.one}
\begin{aligned}
F_{error}(w) =~& P({Y}=+1)  \left (1-\phi \left (\mu_{Z^+}/\sigma_{Z^+}\right ) \right ) \\
&+ P({Y}=-1)  \phi \left (\mu_{Z^-}/\sigma_{Z^-}\right ),
\end{aligned}
\end{equation}
where $\mu_{Z^+} = w^T \mu^+$, $\sigma_{Z^+} = \sqrt{w^T \Sigma^+ w}$, $\mu_{Z^-} = w^T \mu^-$, and $\sigma_{Z^-} = \sqrt{w^T \Sigma^- w}$, and $\phi$ is the CDF of the standard normal distribution, so that $\phi(x) = \int_{-\infty}^x\frac{1}{\sqrt{2 \pi}}\exp({-\frac{1}{2} t^2}) dt$, for $\forall x \in \mathbb{R}$.
\end{theorem} 


In Theorem \ref{col.error.short} we show that the explicit derivative of $F_{error}(w)$ over $w$ can be obtained. To this end, first we need to state the first derivative of the cumulative function $\phi\left (f(w)\right )$, where $f(w) = {w^T \hat{\mu}}/{\sqrt{w^T \hat{\Sigma} w}}$, as is summarized in Lemma \ref{lemma.derivative.phi}.

\begin{lemma} \label{lemma.derivative.phi}
The first derivative of the cumulative function 
\begin{equation*}
\begin{aligned}
&\phi\left (g(w)\right ) =  \int_{-\infty}^{g(w)} \frac{1}{\sqrt{2 \pi}}\exp\left ({-\frac{1}{2} t^2}\right ) dt,\\
\end{aligned}
\end{equation*}
\text{with}~~$g(w) = \frac{w^T \hat{\mu}}{\sqrt{w^T \hat{\Sigma} w}}$ is
\begin{equation*}
\begin{aligned}
\frac{d}{dw} \phi(g(w)) & = \frac{1}{\sqrt{2 \pi}} \exp\left ({-\frac{1}{2} \left ({\frac{w^T \hat{\mu}}{\sqrt{w^T \hat{\Sigma} w}}}\right )^2}\right ) \\
&  \left ( \frac{\sqrt{w^T \hat{\Sigma} w} \cdot \hat{\mu} - {\frac{w^T \hat{\mu}}{\sqrt{w^T \hat{\Sigma} w}}} \cdot \hat{\Sigma} w}{w^T \hat{\Sigma} w} \right ).
\end{aligned}
\end{equation*}
\end{lemma} 
%
%


\begin{theorem} \label{col.error.short}
The derivative of the smooth function $F_{error}(w)$ is defined as
\begin{equation*}
\begin{aligned}
&\nabla_w F_{error}(w) \\
=~& P\left (Y = -1\right ) \frac{1}{\sqrt{2 \pi}} \exp\left (-\frac{1}{2} \left (\frac{\mu_{Z^-}}{\sigma_{Z^-}}\right )^2\right ) \\
 & \cdot \left ( \frac{\sigma_{Z^-}{\mu}^- -  \frac{\mu_{Z^-}}{\sigma_{Z^-}}  \cdot \Sigma^- w}{\sigma_{Z^-}^2} \right ) \\
&- P\left (Y = +1\right ) \frac{1}{\sqrt{2 \pi}} \exp\left (-\frac{1}{2} \left (\frac{\mu_{Z^+}}{\sigma_{Z^+}}\right )^2\right ) \\
&\cdot  \left ( \frac{\sigma_{Z^+} {\mu}^+ -  \frac{\mu_{Z^+}}{\sigma_{Z^+}} \cdot  \Sigma^+ w}{\sigma_{Z^+}^2} \right ),
\end{aligned}
\end{equation*}
where $\mu_{Z^-}$, $\sigma_{Z^-}$, $\mu_{Z^+}$, and $\sigma_{Z^+}$ are defined in Theorem \ref{theorem.main.0_1}.
\end{theorem}

For the rest of this section, we show that $F_{AUC}(w)$  is a smooth function and derive its closed form expression under the Gaussian assumption. 
First, let us restate \eqref{def.AUC} using probabilistic interpretation, e.g.,

\begin{equation}\label{F_AUC}
\begin{aligned}
F_{AUC}(w) &=1- AUC(f)\\
 &= 1-\mathbb{E}_{\mathcal{X}^+,\mathcal{X}^-} \left [\mathbbm{1}\left [f\left (X^+;w\right )> f\left (X^-;w\right )\right ] \right ]  \\
 &= 1 - P\left (w^T {X}^+ >w^T {X}^-\right ) \\
& = 1 - P\left (w^T \left (X^- - X^+\right ) < 0 \right ).
\end{aligned}
\end{equation}

As in the case of $F_{error}(w)$, the smoothness of $F_{AUC}(w)$ follows from the smoothness of the CDF of  $w^T \left (X^- - X^+\right ) $. We will also  use \emph{Corollary 3.3.1} from \cite{tong}, stated as what follows.

\begin{theorem} \label{T1}
If  two $d-$dimensional random vectors ${X}^+$ and ${X}^-$ have a joint multivariate Gaussian distribution, such that
\begin{equation} \label{p1}
\begin{pmatrix} {X}^+ \\ {X}^- \end{pmatrix}
\sim \mathcal{N}\left( \mu, \Sigma \right),
\end{equation}
\begin{equation*}
 \text{where} ~~ \mu = \begin{pmatrix} \mu^+ \\ \mu^- \end{pmatrix}~~\text{and}~~~ \Sigma = 
\begin{pmatrix} \Sigma^{++} & \Sigma^{+-} \\ \Sigma^{-+} & \Sigma^{--} \end{pmatrix}.
\end{equation*}
Then, the marginal distributions of ${X}^+$ and ${X}^-$ are normal distributions with the following properties
\begin{equation*}
{X}^+ \sim \mathcal{N}\left(\mu^+, \Sigma^{++}\right)~~~ \text{and}~~~{X}^- \sim \mathcal{N}\left(\mu^-, \Sigma^{--}\right).
\end{equation*}
\end{theorem}

Further, we need to use \emph{Corollary 3.3.3} in \cite{tong}.

\begin{theorem} \label{T2}
Consider two random vectors ${X}^+$ and ${X}^-$, as defined in \eqref{p1}, then for any vector $w \in \mathbb{R}^d$, we have
\begin{equation} \label{z}
Z = w^T \left({X}^--{X}^+\right) \sim \mathcal{N}\left(\mu_Z, \sigma_Z\right),~~~\text{where}
\end{equation}
\begin{equation} \label{zp}
\begin{aligned}
 \mu_Z &= w^T\left(\mu^- - \mu^+\right)~~\text{and}~~\\
\sigma_Z &= \sqrt{w^T \left( \Sigma^{--} + \Sigma^{++} - \Sigma^{-+} - \Sigma^{+-} \right)w}.
\end{aligned}
\end{equation}
\end{theorem}

Now, in what follows, we derive the formula for $F_{AUC}(w)$ under the Gaussian assumption. 
\begin{theorem} \label{T3}
If two random vectors ${X}^+$ and ${X}^-$ have a joint normal distribution as is defined in Theorem \ref{T1}, then we have
\begin{equation} \label{F.AUC.smooth}
F_{AUC}(w) = 1 - \phi\left(\frac{\mu_Z}{\sigma_Z}\right),
\end{equation}
where $\phi$ is the CDF of the standard normal distribution, so that $\phi(x) = \int_{-\infty}^x\frac{1}{\sqrt{2 \pi}}\exp({-\frac{1}{2} t^2}) dt$, for $\forall x \in \mathbb{R}$ and $\mu_{Z}$ and $\sigma_{Z}$ are defined in \eqref{zp}.
\end{theorem} 
%

In Theorem \ref{T3}, since the CDF of the standard normal distribution $\phi(\cdot)$ is a smooth function, we can conclude that for linear classifiers, the corresponding $F_{AUC}(w)$ is a smooth function of $w$. Moreover, it is possible to compute the derivative of this function, if the first and second moments of the normal distribution are known, as is stated in the following theorem. 

\begin{theorem} \label{col.AUC}
The derivative of the smooth function $F_{AUC}(w)$ is defined as

\begin{equation*}
\begin{aligned}
&\nabla_w F_{AUC}(w) \\=~&
 - \frac{1}{\sqrt{2 \pi}} \exp\left (-\frac{1}{2} 
\left (\frac{\mu_Z}{\sigma_Z}\right )^2\right )\left 
( \frac{\sigma_Z \cdot \hat{\mu} -  {\frac{\mu_Z}{\sigma_Z}}
\cdot {\hat{\Sigma}} w}{\sigma_Z^2} \right ).
\end{aligned}
\end{equation*}
where $\hat{\mu} = \mu^- - \mu^+$ and $\hat{\Sigma} = \Sigma^{--} + \Sigma^{++} - \Sigma^{-+} - \Sigma^{+-}$, and $\mu_Z$ and $\sigma_Z$ are defined in \eqref{zp}.
\end{theorem}


In the next section, we will apply the classical  \emph{gradient descent with backtracking line search} to optimize the expected risk and the expected AUC directly and compare the results of this optimization to optimizing  $F_{log}(w)$ and $F_{hinge}(w)$, respectively. We apply our method to standard data sets for which Gaussian assumption may not hold. It is important to note that our proposed method relies on the assumption that $w^T X$  and $w^T \left (X^- - X^+\right ) $ are Gaussian random variables with moments that are derived from the moments of the original distribution of $X$. In \cite{centralLimit}, it is shown that the distribution of the sums of partially dependent random variables approach  normal distribution under some conditions of the dependency.
Based on these results we believe that while the data itself may not be Gaussian, the random variables  $w^T X $ and  $w^T \left (X^- - X^+\right ) $ may have a nearly normal distribution whose  CDF is well approximated by the CDF in Theorems \ref{theorem.main.0_1} and \ref{T2}, respectively.  
To support our observation further, we compared the linear classifiers obtained by our proposed methods to those obtained by LDA which is a well-known method to produce linear classifiers under the Gaussian assumption. We observed that the accuracy obtained by the LDA classifiers is significantly worse than that of obtained by either our approach or by optimizing surrogate loss function. Hence, we conclude that the behavior of our proposed approach is not strongly dependent on the original Gaussian assumption. Theoretical justification of this claim is a subject for the future research.

\section{Numerical Analysis}\label{sec.numerical}
First we compare the performance of the linear classifiers  obtained by directly optimizing the expected risk versus those obtained by  regularized logistic regression. We use  gradient descent as is stated in Algorithm \ref{GD}.


\begin{algorithm}[ht]
\caption{~\textbf{Gradient Descent with Backtracking Line Search}}
\label{GD}
\begin{algorithmic}
\STATE \text{1:} Initialize $w_0 \in \mathbb{R}^d$, and choose $c \in (0,1)$, and $\beta \in (0,1)$. 
\STATE \textbf{2:} \textbf{for $i = 1,2, \cdots$ do}
\STATE \text{3:}~~~~Choose $\alpha_k^0$ and define $\alpha_k := \alpha_k^0$.
\STATE \text{4:}~~~~Compute the trial point 
\begin{equation*}
w_k^{trial} \gets w_{k-1} - \alpha_k \nabla_w F(w_k).
\end{equation*}
\STATE \textbf{5:} ~~~~\textbf{while $F(w_k^{trial}) > F(w_k) + c \alpha_k \|\nabla_w F(w_k)\|^2$ do}
\STATE \textbf{6:} ~~~~~~~~Set $\alpha_k \gets \beta \alpha_k$.
\STATE \textbf{7:} ~~~~~~~~Compute $w_k^{trial} \gets w_{k-1} - \alpha_k \nabla_w F(w_k)$.
\STATE \textbf{8:} ~~~~\textbf{end while.}
\STATE \textbf{8:} \textbf{end for.}
 \end {algorithmic}
\end{algorithm}
 We perform Algorithm \ref{GD} to  $F(w) = F_{error}(w)$ defined in \eqref{smooth.zero.one},  and to $F(w) = F_{log}(w)$ defined in \eqref{min.logreg}. 

 $F_{error}(w)$ is a nonconvex function, thus in an attempt to avoid bad local minima  we generate a starting point as follows
\begin{equation*}
w_0=\frac{\bar w_0}{\|\bar w_0\|}, \quad \text{where}\quad \bar w_0 = \mu^+ - \frac{{\mu^-}^T \mu^+}{\|\mu^-\|^2} \mu^-.
\end{equation*}

We set the parameters of Algorithm \ref{GD} as $c = 10^{-4}$, $\beta = 0.5$, and $\sigma_k^0 = 1$ and terminate the algorithm 
when $\|\nabla_w F(w_k) \| < 10^{-7} \|\nabla_w F(w_0) \|$ or when the maximum number of iterations 250 is reached. 
For the  logistic regression, the regularization parameter in \eqref{min.logreg} is set as $\lambda = 1/n$, and the initial point 
$w_0$ is selected randomly, since the optimization problem is convex. 

All experiments, implemented in Python 2.7.11, were performed on a computational cluster consisting of 16-cores AMD Operation, 2.0 GHz nodes with 32 Gb of memory.

We considered artificial data sets generated from normal distribution and real data sets.  We have generated 9 different artificial Gaussian data sets of various dimensions using random first and second moments,  summarized in Table \ref{t1_data_art}. Moreover, we generated data sets with some percentage of outliers by swapping a  specified percentage of  positive and negative examples.

\begin{table}[H]  
\centering
\caption{Artificial data sets statistics. $d:$ number of features, $n:$ number of data points, $P^+, P^-:$ prior probabilities, \\$out:$ percentage of outlier data points.}
\label{t1_data_art}
\begin{center}
\begin{small}
 \begin{tabular}{ccccccr} 
Name&$d$&${n}$&$P^+$&$P^-$&$out \%$\\ 
\hline
$data_1$&500&5000&0.05&0.95&0\\
$data_2$&500&5000&0.35&0.65&5\\
$data_3$&500&5000&0.5&0.5&10\\
$data_4$&1000&5000&0.15&0.85&0\\
$data_5$&1000&5000&0.4&0.6&5\\
$data_6$&1000&5000&0.5&0.5&10\\
$data_7$&2500&5000&0.1&0.9&0\\
$data_8$&2500&5000&0.35&0.65&5\\
$data_9$&2500&5000&0.5&0.5&10\\
\end{tabular}
\end{small}
\end{center}
\vskip -0.1in
\end{table}

The corresponding numerical results are summarized in Table \ref{t_art_0-1}, where we used 80 percent of the data points as the training data and the rest  as the test data. The reported average accuracy is based on 20 runs for each data set.  When minimizing $F_{error}(w)$, we  used the exact moments from which the data set was  generated, and also the approximate moments, empirically obtained through the sampled data points. 

We see in Table \ref{t_art_0-1} that, as expected, minimizing $F_{error}(w)$ using the exact moments produces linear classifiers with superior performance overall, while  minimizing $F_{error}(w)$ using approximate moments outperforms  minimizing $F_{log}(w)$. In Table \ref{t_art_0-1}, the bold numbers indicate the 
average testing accuracy attained by  minimizing $F_{error}(w)$ using approximate moments, when this accuracy is significantly better than that obtained by minimizing $F_{log}(w)$. Note also that minimizing $F_{error}(w)$ requires less time than minimizing $F_{log}(w)$.  

\begin{table}[H]   
\centering
 \caption{$F_{error}(w)$ vs. $F_{log}(w)$ minimization via Algorithm \ref{GD} on artificial data sets.}
   \vskip 0.1in
  \label{t_art_0-1}
   \vskip 0.05in
\begin{center}
\begin{small}
 \begin{tabular}{ p{0.6cm}|p{0.9cm}p{0.6cm}|p{0.9cm}p{0.9cm}|p{0.9cm}p{0.6cm}}  
\multirow{4}{*}{Data} 
&\multicolumn{2}{c|}{\textbf{$\pmb{F_{error}(w)}$ Min. }}&\multicolumn{2}{c|}{\textbf{$\pmb{F_{error}(w)}$ Min. }}&\multicolumn{2}{c}{\textbf{$\pmb{F_{log}(w)}$ Min.}}\\  
&\multicolumn{2}{c|}{\textbf{Exact}}&\multicolumn{2}{c|}{\textbf{Approximate}}&&\\ 
&\multicolumn{2}{c|}{\textbf{moments}}&\multicolumn{2}{c|}{\textbf{moments}}&&\\ 
&{Accuracy}&{Time(s)}&{Accuracy}&{Time(s)}&{Accuracy}&{Time(s)}\\
\hline
$data_1$&0.9965&0.25&0.9907&1.04&0.9897&3.86\\ 
$data_2$&0.9905&0.26&\textbf{0.9806}&0.86&0.9557&13.72\\
$data_3$&0.9884&0.03&\textbf{0.9745}&1.28&0.9537&15.79\\ 
$data_4$&0.9935&0.63&0.9791&5.51&0.9782&10.03\\
$data_5$&0.9899&5.68&\textbf{0.9716}&10.86&0.9424&28.29\\
$data_6$&0.9904&0.83&\textbf{0.9670}&5.18&0.9291&25.47\\
$data_7$&0.9945&4.79&0.9786&32.75&0.9697&43.20\\ 
$data_8$&0.9901&9.96&0.9290&119.64&0.9263&104.94\\
$data_9$&0.9899&1.02&0.9249&68.91&0.9264&123.85\\
\end{tabular}
\end{small}
\end{center}
\vskip -0.1in
\end{table}

Further, we used 19 real data sets downloaded from LIBSVM website\footnote{\url{https://www.csie.ntu.edu.tw/~cjlin/libsvmtools/datasets/binary.html}} and UCI machine learning repository\footnote{\url{http://archive.ics.uci.edu/ml/}}, summarized in Table \ref{t1}. We have normalized  the data sets so that each dimension has mean 0 and variance 1.  The data sets from UCI machine learning repository with categorical features are transformed into grouped binary features.

\begin{table}[H]  
\centering
\caption{Real data sets statistics. $d:$ number of features, $n:$ number of data points, $P^+, P^-:$ prior probabilities, \\$AC:$ attribute characteristics.}
\label{t1}
\begin{center}
\begin{small}
 \begin{tabular}{ccccccr} 
Name&AC&$d$&${n}$&$P^+$&$P^-$\\ 
\hline
fourclass&real&2&862&0.35&0.65\\
svmguide1&real&4&3089&0.35&0.65\\
diabetes&real&8&768&0.35&0.65\\
shuttle&real&9&43500&0.22&0.78\\
vowel&int&10&528&0.09&0.91\\
magic04&real&10&19020&0.35&0.65\\
poker&int&11&25010&0.02&0.98\\
letter&int&16&20000&0.04&0.96\\ 
segment&real&19&210&0.14&0.86\\
svmguide3&real&22&1243&0.23&0.77\\
ijcnn1&real&22&35000&0.1&0.9\\
german&real&24&1000&0.3&0.7\\
landsat satellite&int&36&4435&0.09&0.91\\
sonar&real&60&208&0.5&0.5\\
a9a&binary&123&32561&0.24&0.76\\
w8a&binary&300&49749&0.02&0.98\\
mnist&real&782&100000&0.1&0.9\\
colon-cancer&real&2000&62&0.35&0.65\\
gisette&real&5000&6000&0.49&0.51\\
\end{tabular}
\end{small}
\end{center}
\vskip -0.1in
\end{table}

Table \ref{t_0-1} summarizes the  performance comparison between the linear classifiers obtained by  minimizing $F_{error}(w)$ versus $F_{log}(w)$. 
We used \emph{five-fold cross-validation} 
and  repeated each experiment four times, and the average test accuracy over the 20 runs are reported for each problem.

\begin{table}[H]    
\centering
 \caption{$F_{error}(w)$ vs. $F_{log}(w)$ minimization via Algorithm \ref{GD} on real data sets.}
   \vskip 0.1in
  \label{t_0-1}
   \vskip 0.05in
   \begin{center}
\begin{small}
 \begin{tabular}{c|p{1cm}c|p{1cm}c}  
\multirow{2}{*}{Data} 
&\multicolumn{2}{c|}{\textbf{$\pmb{F_{error}(w)}$}}&\multicolumn{2}{c}{\textbf{$\pmb{F_{log}(w)}$}}\\   
&{Accuracy}&{Time (s)}&{Accuracy }&{Time (s)}\\
\hline
fourclass&0.8782&0.02&0.8800&0.12\\ 
svmguide1&\textbf{0.9735}&0.42&0.9506&0.28\\ 
diabetes&0.8832&1.04&0.8839&0.13\\ 
shuttle&0.8920&0.01&\textbf{0.9301}&4.05\\ 
vowel&0.9809&0.91&0.9826&0.11\\ 
magic04&0.8867&0.66&0.8925&1.75\\  
poker&0.9897&0.17&0.9897&10.96\\ 
letter&0.9816&0.01&0.9894&4.51\\ 
segment&0.9316&0.17&\textbf{0.9915}&0.36\\ 
svmguide3&\textbf{0.9118}&0.39&0.8951&0.17\\ 
ijcnn1&0.9512&0.01&0.9518&4.90\\  
german&0.8780&1.09&0.8826&0.62\\ 
landsat satellite&0.9532&0.01&0.9501&3.30\\ 
sonar&\textbf{0.8926}&0.49&0.8774&0.92\\ 
a9a&0.9193&0.98&0.9233&11.45\\ 
w8a&0.9851&0.36&0.9876&24.16\\ 
mnist&0.9909&3.79&0.9938&136.83\\ 
colon cancer&\textbf{0.9364}&15.92&0.8646&1.20\\ 
gisette&0.9782&310.72&0.9706&136.71\\
\end{tabular}
\end{small}
\end{center}
\vskip -0.1in
\end{table}

As we can see in Table \ref{t_0-1}, the linear classifier obtained by minimizing $F_{error}(w)$ has comparable test accuracy to the one obtained from minimizing $F_{log}(w)$ in 13 cases out of 19. In 4 cases minimizing $F_{error}(w)$ surpasses minimizing $F_{log}(w)$ in terms of the average test accuracy, while performs worse in the case of the 2 remaining data sets. Finally,  we note that the solution time of optimizing $F_{error}(w)$ is significantly less than that of optimizing $F_{log}(w)$ when $d$ is smaller than $n$. 

\begin{figure} [H]
\begin{subfigure}
  \centering
  \includegraphics[height=0.35\linewidth,width=0.49\linewidth]{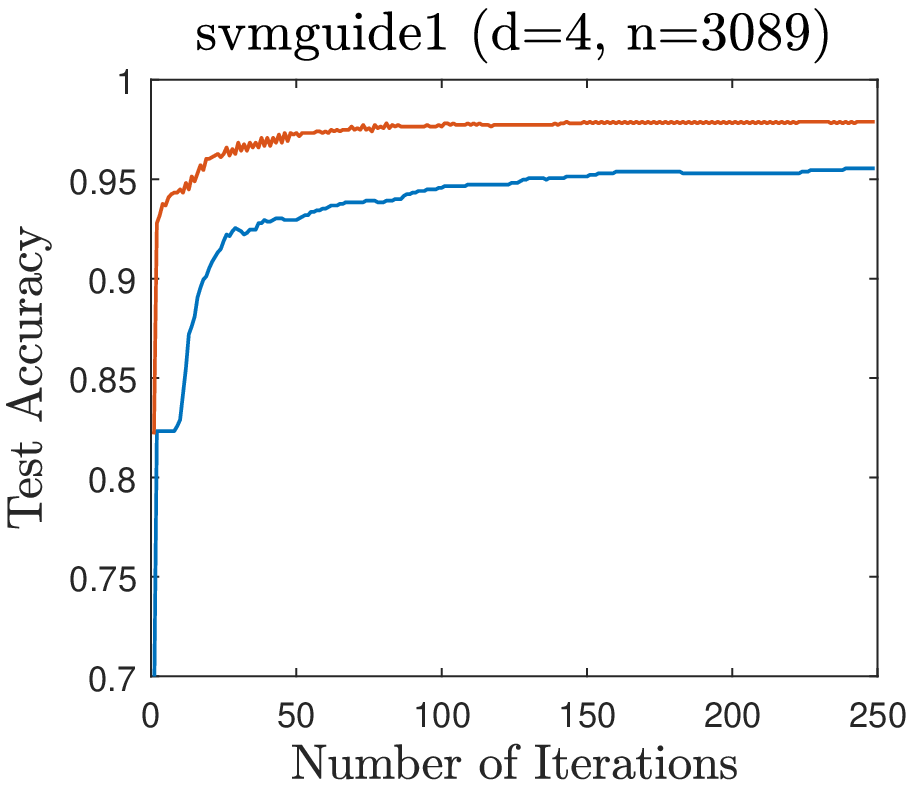}
  \label{fig:sfig1}
\end{subfigure}%
\begin{subfigure}
  \centering
  \includegraphics[height=0.35\linewidth,width=0.5\linewidth]{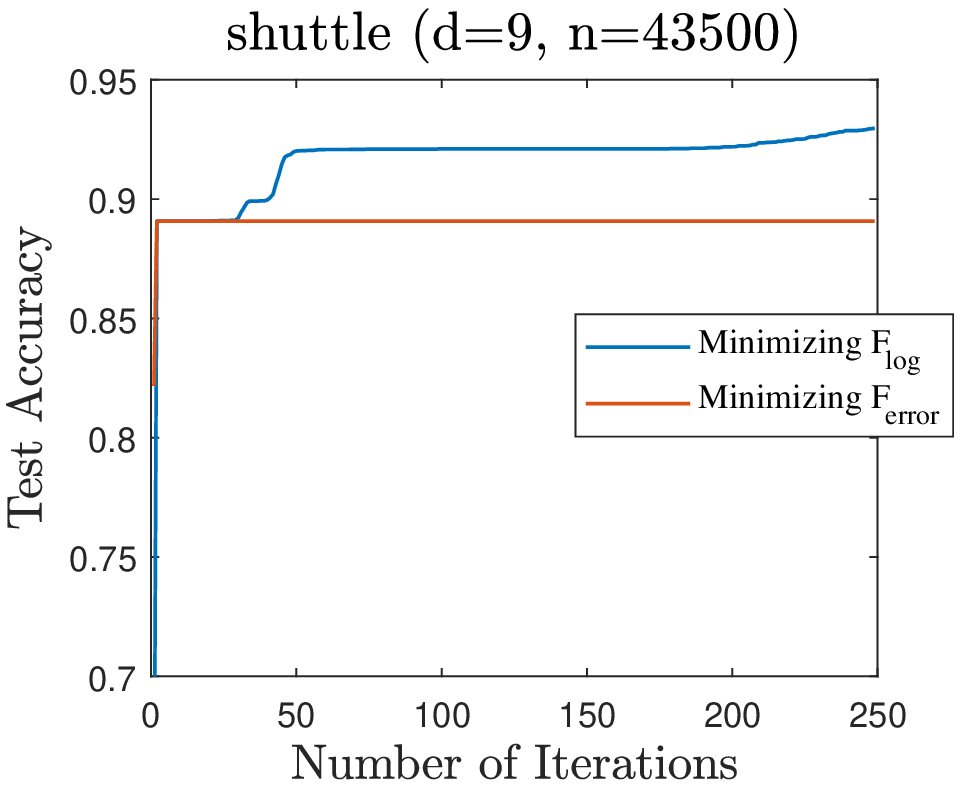}
  \label{fig:sfig1}
\end{subfigure}%
\begin{subfigure}
 \centering
  \includegraphics[height=0.35\linewidth,width=0.49\linewidth]{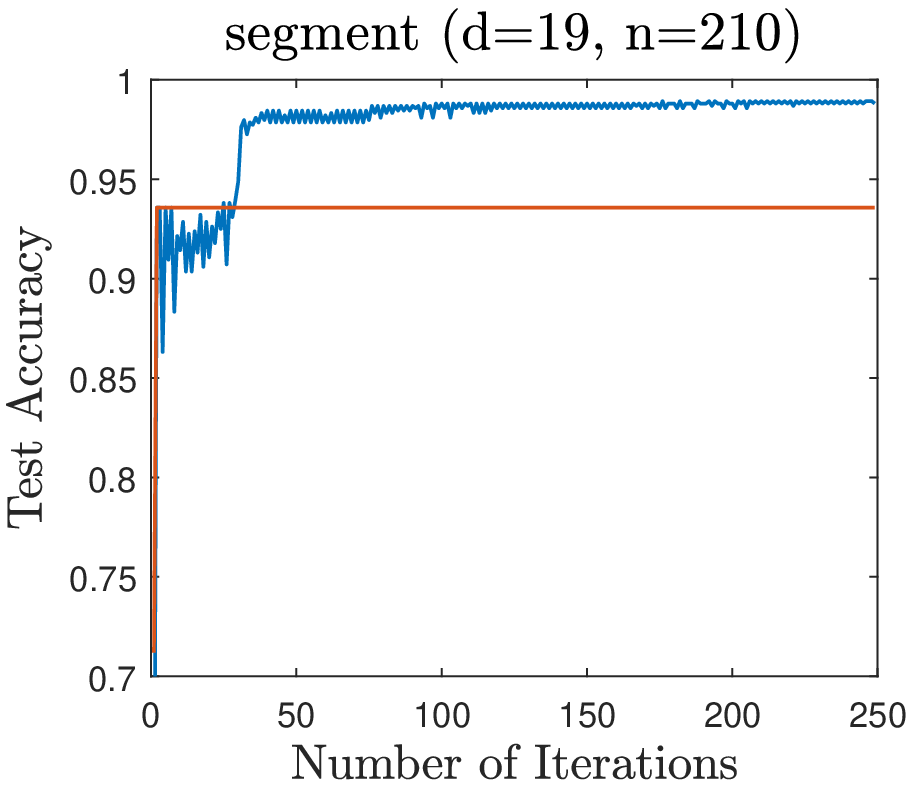}
  \label{fig:sfig5}
\end{subfigure}%
\begin{subfigure}
  \centering
  \includegraphics[height=0.35\linewidth,width=0.49\linewidth]{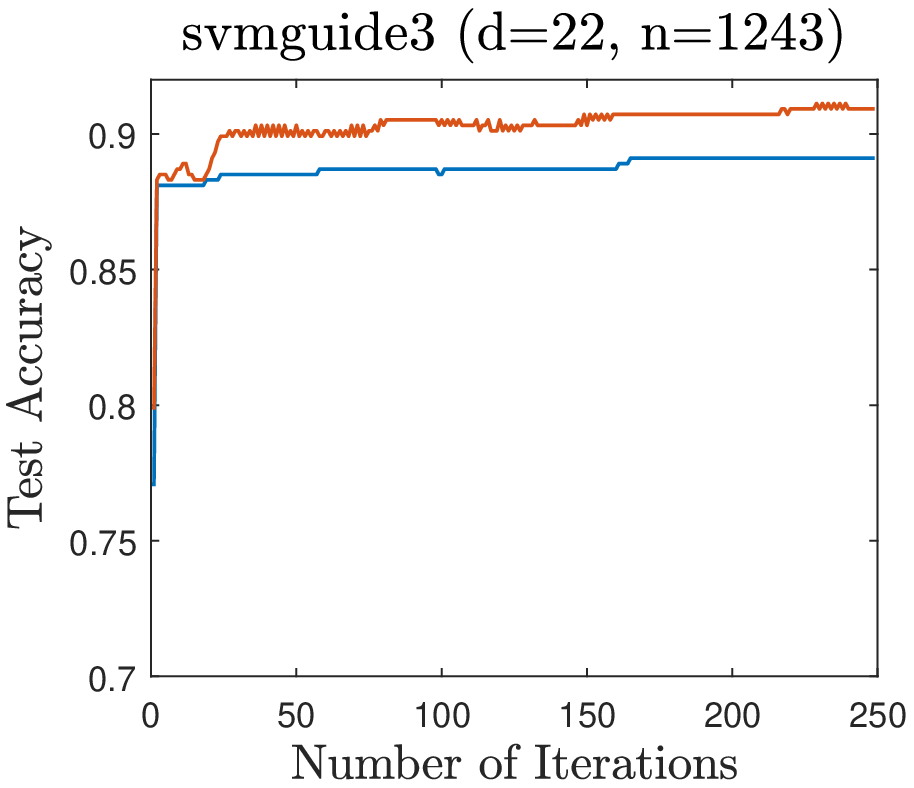}
  \label{fig:sfig6}
\end{subfigure} %
\begin{subfigure}
 \centering
  \includegraphics[height=0.35\linewidth,width=0.49\linewidth]{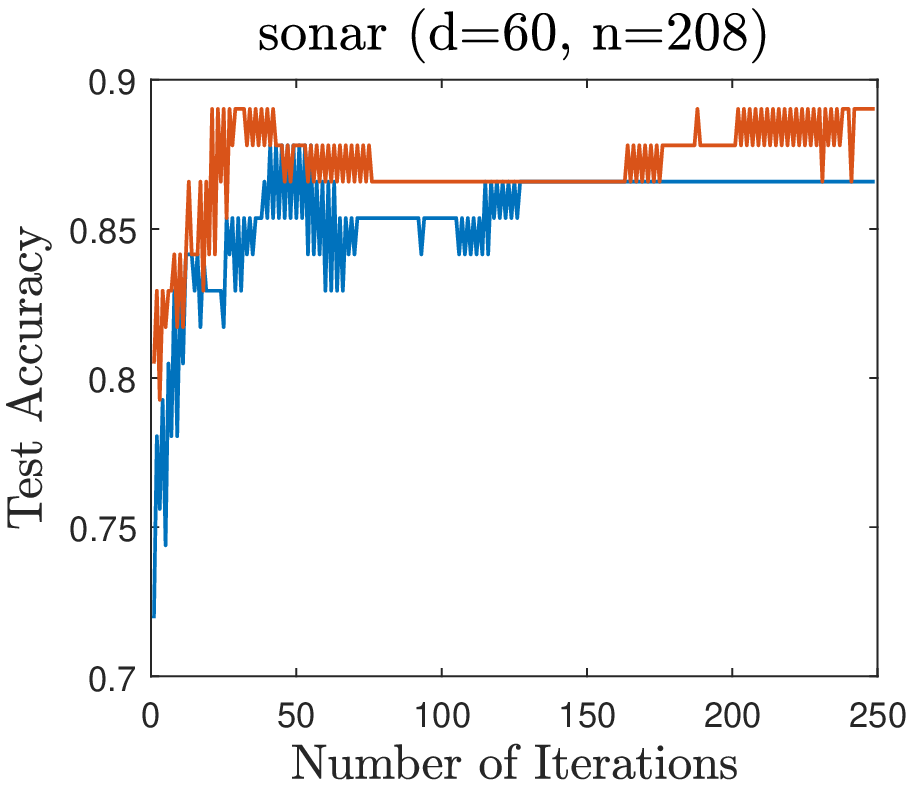}
  \label{fig:sfig5}
\end{subfigure}%
\begin{subfigure}
  \centering
  \includegraphics[height=0.35\linewidth,width=0.49\linewidth]{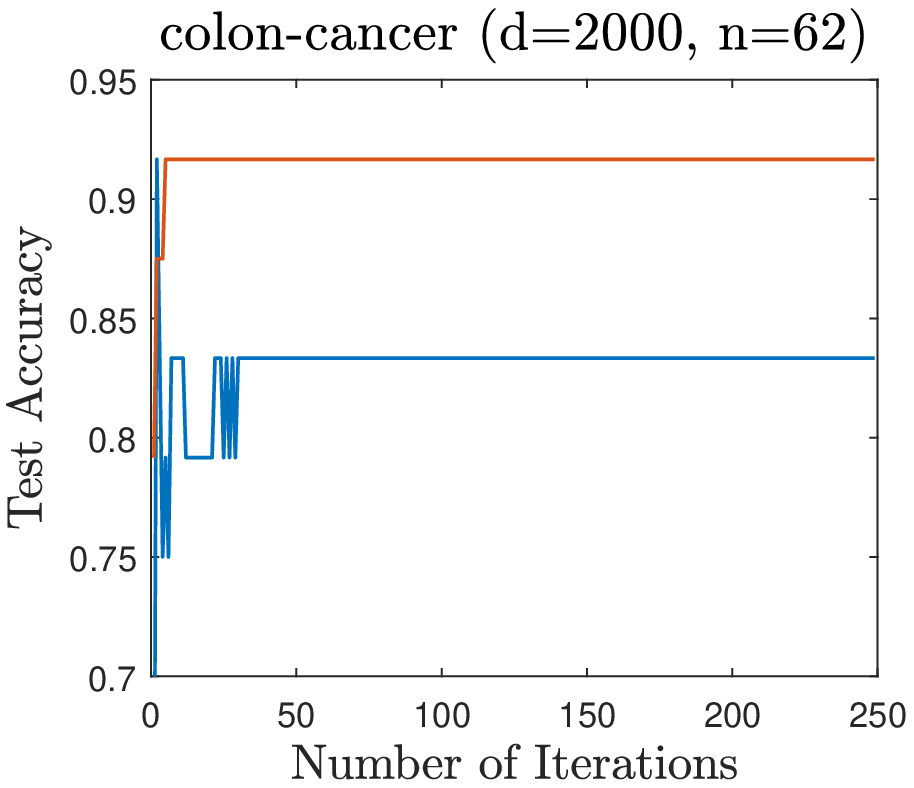}
  \label{fig:sfig6}
\end{subfigure} %

\caption{Performance of minimizing $F_{error}(w)$ vs. $F_{log}(w)$ via Algorithm \ref{GD}.}
\label{fig1}
\end{figure}

Figure \ref{fig1} illustrates the progress of the linear classifiers obtained through these two different approaches in terms of the average test accuracy over iterations. In Figure \ref{fig1} we selected the data sets in which minimizing $F_{error}(w)$ has a better performance in terms of the final test accuracy compared to minimizing $F_{log}(w)$ or vice versa. We note that in two cases where optimizing $F_{error}(w)$ performs worse than minimizing $F_{log}(w)$, the algorithm achieved its best $F_{error}(w)$ value during the first few iterations and then stalled. This may be due to the inaccurate gradient or simply a local minimum. Improving our method for such cases is subject of future work. 

The comparison with LDA can be found in the Appendix.

We now turn to  comparing the performance of  linear classifiers obtained by optimizing the AUC function, e.g., $F(w) =  F_{AUC}(w)$ defined in \eqref{F.AUC.smooth} and its approximation via pairwise hinge loss, e.g., $F(w) = F_{hinge}(w)$ as is defined in \eqref{AUC_hinge}. The setting of the parameters and the type of the artificial and real data sets are the same as  in Tables \ref{t1_data_art} and \ref{t1}.

The  results for artificial data sets are summarized in Table \ref{t_art_AUC} as the same manner of Table \ref{t_art_0-1}, except that we report the AUC value as the performance measure of the resulting classifiers. As we can see in Table \ref{t_art_AUC}, in the process of minimizing $F_{AUC}(w)$, the only advantage of using the exact moments rather than the approximate moments is in terms of the solution time, since both approaches result in comparable average AUC values. On the other hand, the performance of the linear classifier obtained through minimizing $F_{AUC}(w)$ using approximate moments surpasses that of the classifier obtained via minimizing $F_{hinge}(w)$, both in terms of the average AUC value as well as the required solution time.

\newpage 
\begin{table}[H]     
\centering 
 \caption{$F_{AUC}(w)$ vs. $F_{hinge}(w)$ minimization via Algorithm \ref{GD} on artificial data sets.}
   \vskip 0.1in
  \label{t_art_AUC}
   \vskip 0.05in
\begin{center}
\begin{small}
 \begin{tabular}{ p{0.6cm}|p{0.7cm}p{0.6cm}|p{0.7cm}p{0.9cm}|p{0.7cm}p{0.6cm}}  
\multirow{4}{*}{Data} 
&\multicolumn{2}{c|}{\textbf{$\pmb{F_{AUC}(w)}$ Min.}}&\multicolumn{2}{c|}{\textbf{$\pmb{F_{AUC}(w)}$ Min.}}&\multicolumn{2}{c}{\textbf{$\pmb{F_{hinge}(w)}$ Min.}}\\  
&\multicolumn{2}{c|}{\textbf{Exact}}&\multicolumn{2}{c|}{\textbf{Approximate}}&\\ 
&\multicolumn{2}{c|}{\textbf{moments}}&\multicolumn{2}{c|}{\textbf{moments}}&&\\ 
&{AUC}&{Time(s)}&{AUC}&{Time(s)}&{AUC}&{Time(s)}\\  
\hline
$data_1$&0.9972&0.01&\textbf{0.9941}&0.23&0.9790&5.39\\
$data_2$&0.9963&0.01&\textbf{0.9956}&0.22&0.9634&159.23\\
$data_3$&0.9965&0.01&\textbf{0.9959}&0.24&0.9766&317.44\\
$data_4$&0.9957&0.02&\textbf{0.9933}&0.83&0.9782&23.36\\
$data_5$&0.9962&0.02&\textbf{0.9951}&0.80&0.9589&110.26\\
$data_6$&0.9962&0.02&\textbf{0.9949}&0.82&0.9470&275.06\\
$data_7$&0.9965&0.08&\textbf{0.9874}&4.61&0.9587&28.31\\ 
$data_8$&0.9966&0.07&\textbf{0.9929}&4.54&0.9514&104.16\\
$data_9$&0.9962&0.08&\textbf{0.9932}&4.54&0.9463&157.62\\
\end{tabular}
\end{small}
\end{center}
\vskip -0.1in
\end{table}

\begin{figure} [H]
\begin{subfigure}
  \centering
  \includegraphics[height=0.35\linewidth,width=0.49\linewidth]{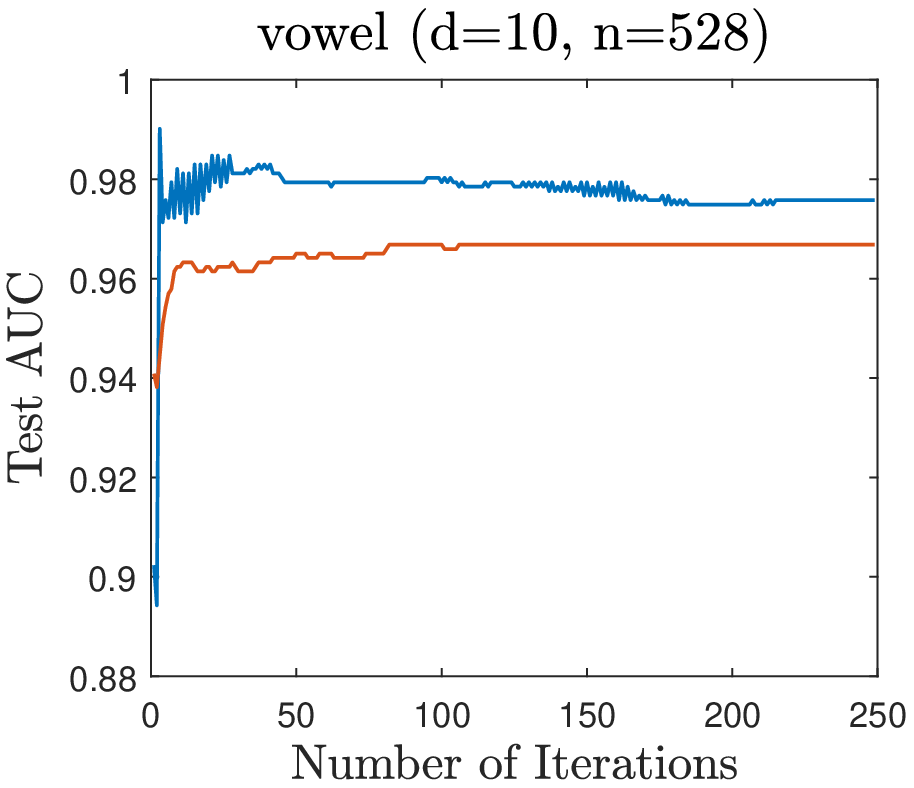}
  \label{fig:sfig1}
\end{subfigure}%
\begin{subfigure}
  \centering
  \includegraphics[height=0.35\linewidth,width=0.5\linewidth]{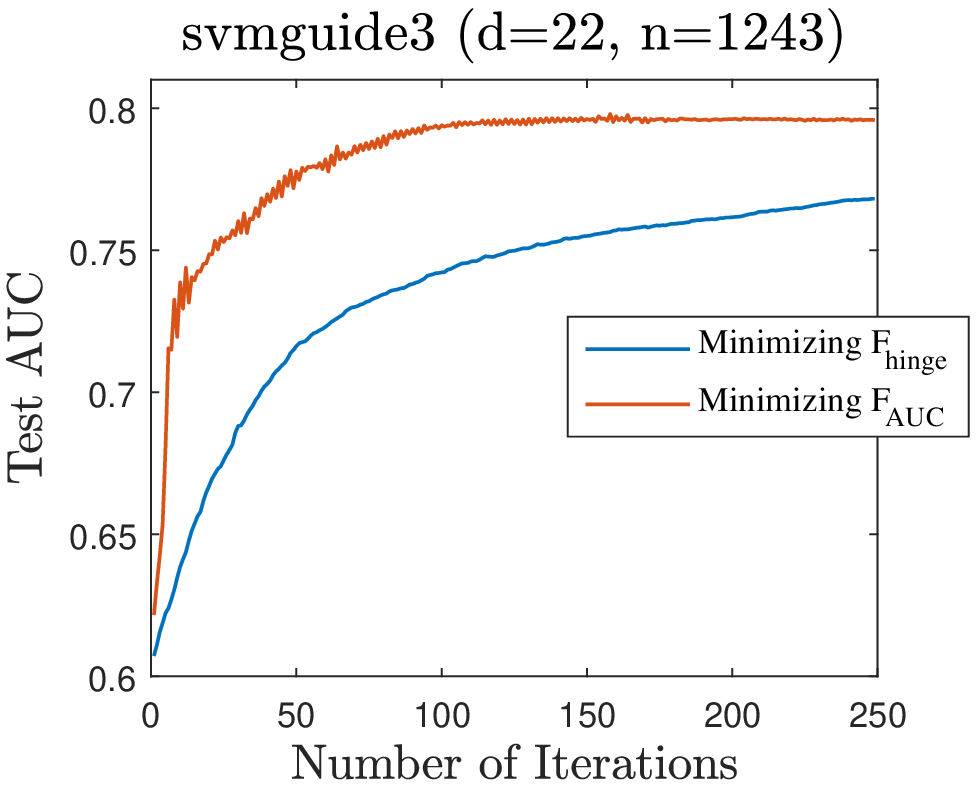}
  \label{fig:sfig1}
\end{subfigure}%
\begin{subfigure}
 \centering
  \includegraphics[height=0.35\linewidth,width=0.49\linewidth]{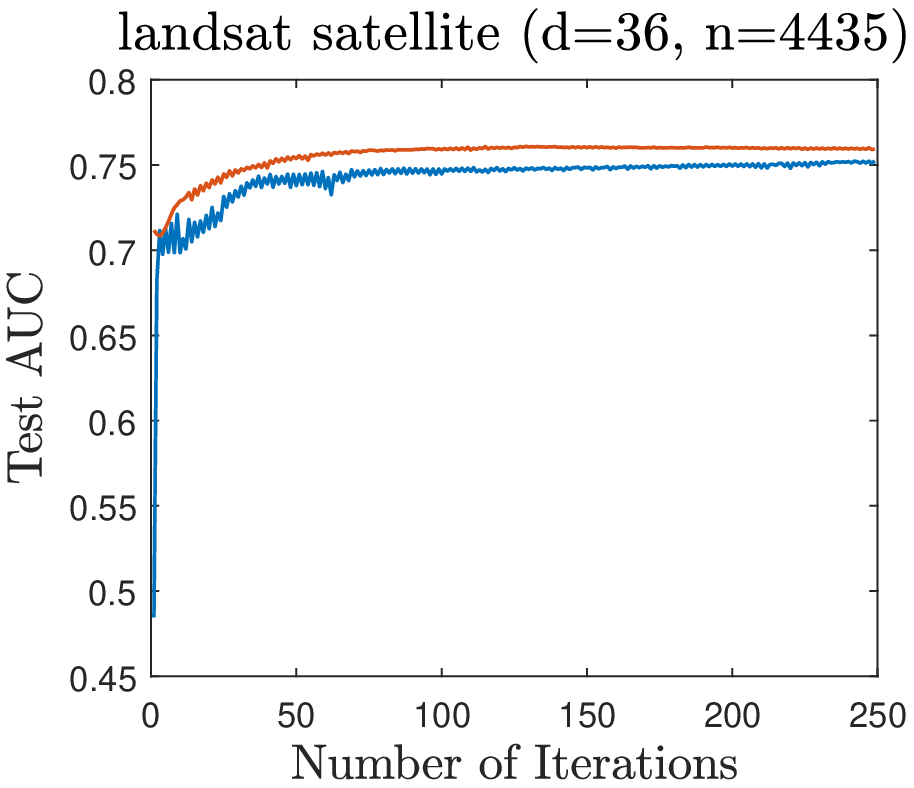}
  \label{fig:sfig5}
\end{subfigure}%
\begin{subfigure}
  \centering
  \includegraphics[height=0.35\linewidth,width=0.49\linewidth]{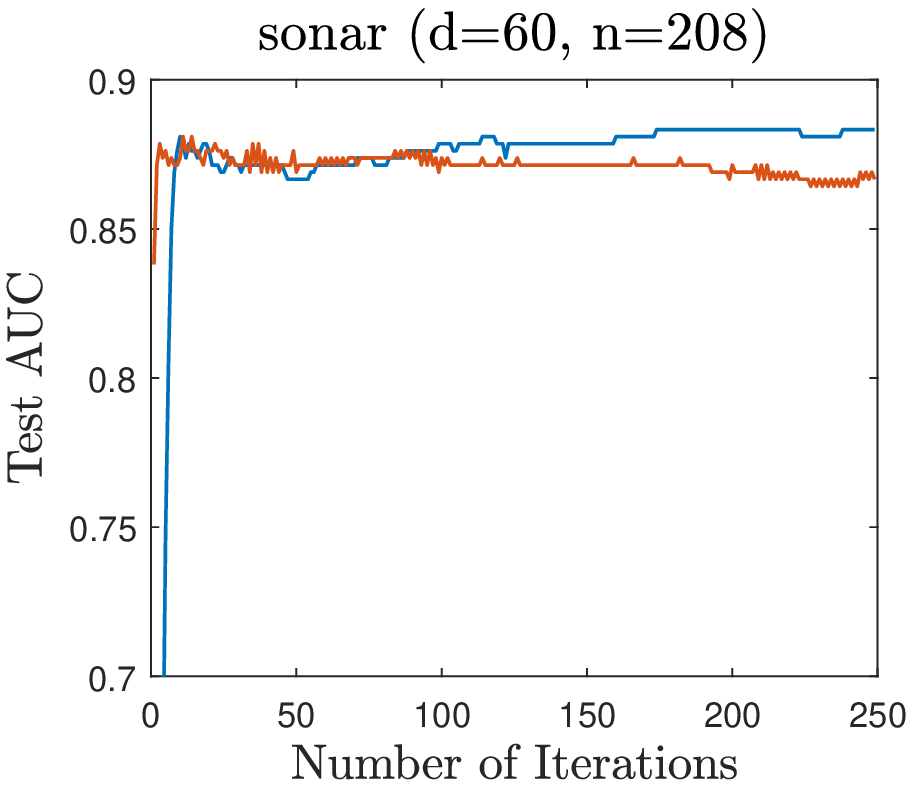}
  \label{fig:sfig6}
\end{subfigure} %
\begin{subfigure}
 \centering
  \includegraphics[height=0.35\linewidth,width=0.49\linewidth]{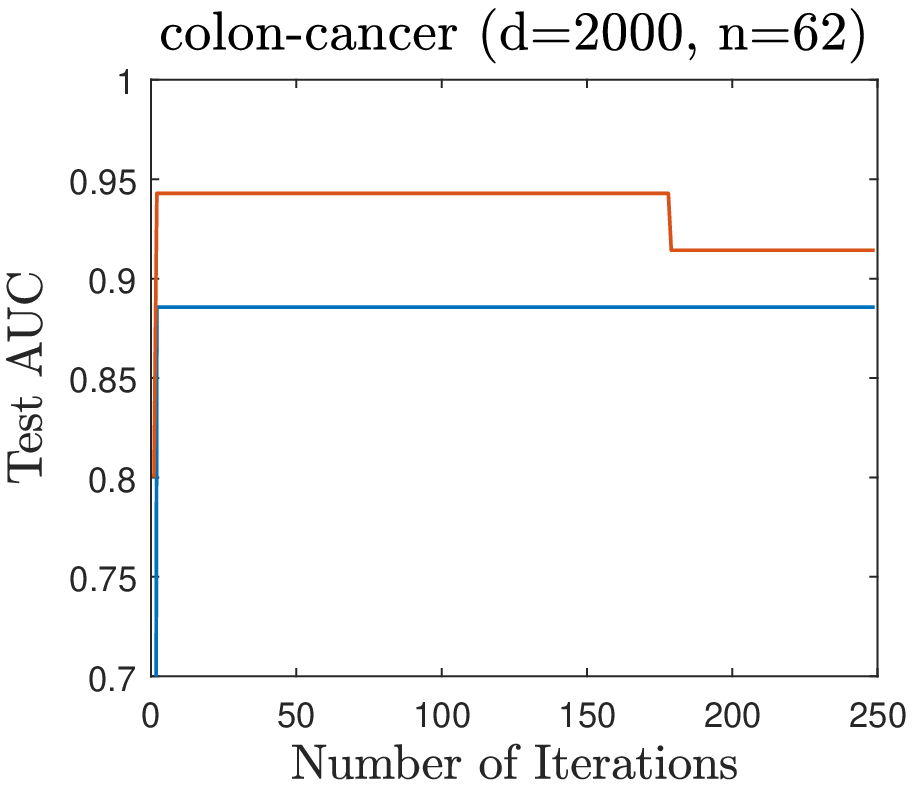}
  \label{fig:sfig5}
\end{subfigure}%
\begin{subfigure}
  \centering
  \includegraphics[height=0.35\linewidth,width=0.49\linewidth]{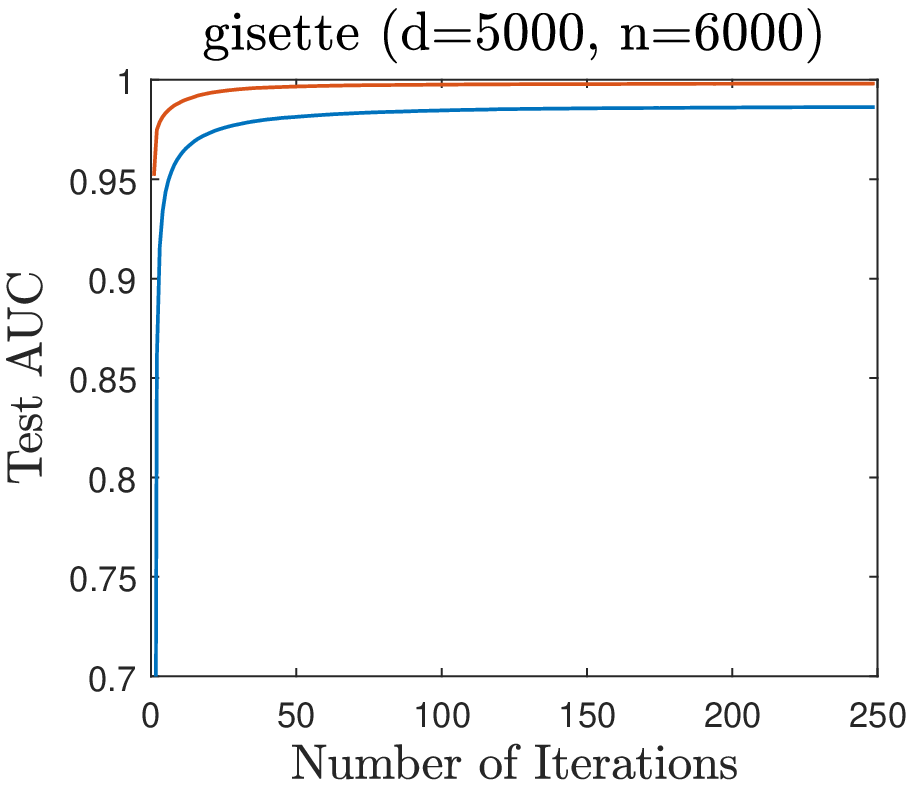}
  \label{fig:sfig6}
\end{subfigure} %

\caption{Performance of minimizing $F_{AUC}(w)$ vs. $F_{hinge}(w)$ via Algorithm \ref{GD}.}
\label{fig2}
\end{figure}

Table \ref{t_AUC} summarizes the results on real data sets, in a  manner similar to Table \ref{t_0-1}, while, again using  AUC of the resulting classifier as the performance measure. As we can see in Table \ref{t_AUC}, the average AUC values of the linear classifiers obtained through minimizing $F_{AUC}(w)$ and $F_{hinge}(w)$ are comparable in 14 cases out of 19, in 4 cases minimizing $F_{AUC}(w)$ performs better than minimizing $F_{hinge}(w)$ in terms of the average test AUC, while their performance is worse in the remaining 2 cases, where the algorithm stalled after a few iterations of optimizing $F_{AUC}(w)$ as is shown in Figure \ref{fig2}.  In terms of solution time, minimizing $F_{AUC}(w)$ significantly outperforms minimizing $F_{hinge}(w)$, due to the high per-iteration complexity dependence on $n$ of $F_{hinge}(w)$ minimization.

\begin{table}[H]      
\centering
\caption{$F_{AUC}(w)$ vs. $F_{hinge}(w)$ minimization via Algorithm \ref{GD} on real data sets.}
   \vskip 0.1in
   \label{t_AUC}
   \vskip 0.05in
\begin{center}
\begin{small}
 \begin{tabular}{c|p{0.8cm}c|p{0.8cm}c}  
\multirow{2}{*}{Data} 
&\multicolumn{2}{c|}{\textbf{$\pmb{F_{AUC}(w)}$ Min.}}&\multicolumn{2}{c}{\textbf{$\pmb{F_{hinge}(w)}$ Min.}}\\   
&{AUC}&{Time(s)}&{AUC}&{Time(s)}\\  
\hline
fourclass&0.8362&0.01&0.8362&6.81\\ 
svmguide1&0.9717&0.06&0.9863&35.09\\ 
diabetes&0.8311&0.03&0.8308&12.48\\ 
shuttle&0.9872&0.11&0.9861&2907.84\\ 
vowel&0.9585&0.12&\textbf{0.9765}&2.64\\ 
magic04&0.8382&0.11&0.8419&1391.29\\  
poker&0.5054&0.11&0.5069&1104.56\\ 
letter&0.9830&0.12&0.9883&121.49\\ 
segment&0.9948&0.21&0.9992&4.23\\ 
svmguide3&\textbf{0.8013}&0.34&0.7877&23.89\\ 
ijcnn1&0.9269&0.08&0.9287&2675.67\\  
german&0.7938&0.14&0.7919&32.63\\ 
landsat satellite&\textbf{0.7587}&0.43&0.7458&193.46\\ 
sonar&0.8214&0.88&\textbf{0.8456}&2.15\\ 
a9a&0.9004&0.92&0.9027&15667.87\\ 
w8a&0.9636&0.54&0.9643&5353.23\\ 
mnist&0.9943&0.64&0.9933&28410.2393\\ 
colon cancer&\textbf{0.8942}&2.50&0.8796&0.05\\ 
gisette&\textbf{0.9957}&31.32&0.9858&3280.38\\ 
\end{tabular}
\end{small}
\end{center}
\vskip -0.1in
\end{table}

More comprehensive numerical results are provided in the appendix, including the standard deviation of the test accuracy and test AUC.

\section{Conclusion}\label{sec.conclusion}
In this work, we showed that under the Gaussian assumption, the expected prediction error and AUC of linear predictors in binary classification  are smooth functions whose  derivatives  can be computed using the first and second moments of the related normal distribution. We then show that empirical moments of real data sets (not necessarily Gaussian) can be utilized to obtain approximate derivatives. This implies that gradient-based optimization approach can be used to optimize the prediction error and AUC function. In this work, for simplicity, we used gradient descent with backtracking line search and we demonstrated the efficiency of directly optimizing prediction error and AUC function compared to their approximations--logistic regression and pairwise hinge loss, respectively. 
The main advantage of these approaches is that the proposed objective functions and their derivatives are independent of the size of the data sets. Clearly more efficient second-order methods can also be utilized for optimizing these functions, which is a subject for the future research.

\clearpage

\nocite{langley00}
\bibliography{references}
\bibliographystyle{icml2017}

\onecolumn
\appendix
\section{Proofs of results in Section} 

\begin{proof} {(Lemma \ref{ER_prob} proof)}
Note that we can split the whole set $\left \{(X,Y)~:~Y \cdot w^TX <0 \right \} \subset \mathcal{X} \times \mathcal{Y}$ into two disjoint sets as the following:
\begin{equation*}
\left \{(X,Y)~:~Y \cdot w^TX <0 \right \} = \left \{(X^+,+1)~:~w^TX^+ <0 \right \} \cup \left \{(X^-,-1)~:~ w^TX^- \geq 0 \right \}.
\end{equation*}
Now, by using  \eqref{expectedRisk_prob} we will have:
\begin{equation*}
\begin{aligned}
F_{error}(w) =~& P\left({Y} \cdot w^T {X}<0\right) \\
=~& P\left({Y} \cdot w^T {X}<0 \cap {Y} = +1 \right) + P\left({Y} \cdot w^T {X}<0 \cap {Y} = -1 \right) \\
=~& P\left({Y} \cdot w^T {X}<0 \mid {Y} = +1 \right) P\left({Y} = +1\right) + P\left({Y} \cdot w^T {X}<0 \mid {Y} = -1 \right) P\left({Y} = -1\right) \\
=~& P\left( w^T {X}^+<0\right) P\left({Y} = +1\right) + P\left(w^T {X}^->0\right) P\left({Y} = -1\right) \\
=~& P\left( w^T {X}^+\leq 0\right) P\left({Y} = +1\right) + \left(1-P\left(w^T {X}^- \leq 0\right)\right) P\left({Y} = -1\right).
\end{aligned}
\end{equation*} 
\end{proof} 

\begin{proof} (Theorem \ref{theorem.main.0_1} proof) Let us define  random variables $Z^+$ and $Z^-$ as the follows
\begin{equation*}
Z^+ = w^T {X}^+,~~\text{and}~~Z^- = w^T {X}^-.
\end{equation*}
From \eqref{original.normal} and using Theorem \ref{th.normal.close} we have
\begin{equation*}
{Z}^+ \sim \mathcal{N}\left(w^T \mu^+, w^T \Sigma^+ w \right)~~\text{and}~~
{Z}^- \sim \mathcal{N}\left(w^T \mu^-, w^T \Sigma^- w \right).
\end{equation*}
Then, by using Lemma \ref{ER_prob} we conclude the following 
\begin{equation*}
\begin{aligned}
F_{error}(w) &= P\left ({Y} \cdot w^T {X}<0\right ) \\
&= \left (1- \phi\left({ \mu_{Z^+}}/{\sigma_{Z^+}}\right)\right ) P\left ({Y} = +1\right ) + \phi\left({\mu_{Z^-}}/{\sigma_{Z^-}}\right) P\left ({Y} = -1\right ).
\end{aligned}
\end{equation*}
where $\mu_{Z^+} = w^T \mu^+$, $\sigma_{Z^+} = \sqrt{w^T \Sigma^+ w}$, $\mu_{Z^-} = w^T \mu^-$, and $\sigma_{Z^-} = \sqrt{w^T \Sigma^- w}$. 
\end{proof}

\begin{proof}
Note that based on the chain rule we have 
\begin{equation}\label{chain.rule}
\frac{d}{dw} \phi \left (f(w)\right ) = \phi'(g(w)) g'(w).
\end{equation}
By substituting 
\begin{equation*}
\phi'(x) = \frac{d}{dx} \int_{-\infty}^{x} \frac{1}{\sqrt{2 \pi}}\exp\left ({-\frac{1}{2} t^2}\right ) dt =  \frac{1}{\sqrt{2 \pi}}\exp\left ({-\frac{1}{2} x^2}\right )~~~\text{and }
\end{equation*}
\begin{equation*}
g'(w) =  \frac{\sqrt{w^T \hat{\Sigma} w} \cdot \hat{\mu} - {\frac{w^T \hat{\mu}}{\sqrt{w^T \hat{\Sigma} w}}} \cdot \hat{\Sigma} w}{w^T \hat{\Sigma} w},
\end{equation*}
in \eqref{chain.rule} we conclude the result.
\end{proof}

\begin{proof} (Theorem \ref{col.error.short} proof)
Theorem \ref{col.error.short} is an immediate corollary of  the result of Lemma \ref{lemma.derivative.phi}.
\end{proof}


\begin{proof} (Theorem \ref{T3} proof)
From \eqref{F_AUC} and Theorem \ref{T2} we have
\begin{equation*}
\begin{aligned}
 F_{AUC}(w) &= 1 - P(w^T \left (X^- - X^+) < 0 \right ) = 1 - P\left(Z \leq 0\right) \\
&= 1 - P\left(\frac{Z - \mu_Z}{\sigma_Z} \leq \frac{ - \mu_Z}{\sigma_Z}\right) = 1 - \phi\left(\frac{\mu_Z}{\sigma_Z}\right),
\end{aligned}
\end{equation*}

where the random variable $Z$ is defined in \eqref{z}, with the stated mean and standard deviation in \eqref{zp}.
\end{proof}

\begin{proof} (Theorem \ref{col.AUC} proof)
Theorem \ref{col.AUC} is an immediate corollary of  the result of Lemma  \ref{lemma.derivative.phi}, and the symmetric property of $\phi(\cdot)$. 
\end{proof}



\section{Numerical Analysis}
The following tables summarize more comprehensive numerical comparison between minimizing $F_{error}(w)$ versus $F_{log}(w)$ and also between minimizing $F_{AUC}(w)$ versus minimizing $F_{hinge}(w)$.
\begin{table}[H]   
\centering
 \caption{$F_{error}(w)$ vs. $F_{log}(w)$ minimization via Algorithm \ref{GD} on artificial data sets.}
   \vskip 0.1in
  \label{t_art_0-1_2}
   \vskip 0.05in
\begin{center}
\begin{small}
 \begin{tabular}{ c|cc|cc|cc}  
\multirow{3}{*}{Data} 
&\multicolumn{2}{c|}{\textbf{$\pmb{F_{error}(w)}$ Minimization}}&\multicolumn{2}{c|}{\textbf{$\pmb{F_{error}(w)}$ Minimization}}&\multicolumn{2}{c}{\textbf{$\pmb{F_{log}(w)}$ Minimization}}\\  
&\multicolumn{2}{c|}{\textbf{Exact moments}}&\multicolumn{2}{c|}{\textbf{Approximate moments}}&&\\ 
&{Accuracy$\pm$ std}&{Time (s)}&{Accuracy $\pm$ std}&{Time (s)}&{Accuracy $\pm$ std}&{Time (s)}\\
\hline
$data_1$&0.9965$\pm$0.0008&0.25&0.9907$\pm$0.0014&1.04&0.9897$\pm$0.0018&3.86\\ 
$data_2$&0.9905$\pm$0.0023&0.26&\textbf{0.9806}$\pm$0.0032&0.86&0.9557$\pm$0.0049&13.72\\
$data_3$&0.9884$\pm$0.0030&0.03&\textbf{0.9745}$\pm$0.0037&1.28&0.9537$\pm$0.0048&15.79\\ 
$data_4$&0.9935$\pm$0.0017&0.63&0.9791$\pm$0.0034&5.51&0.9782$\pm$0.0031&10.03\\
$data_5$&0.9899$\pm$0.0026&5.68&\textbf{0.9716}$\pm$0.0048&10.86&0.9424$\pm$0.0055&28.29\\
$data_6$&0.9904$\pm$0.0017&0.83&\textbf{0.9670}$\pm$0.0058&5.18&0.9291$\pm$0.0076&25.47\\
$data_7$&0.9945$\pm$0.0019&4.79&0.9786$\pm$0.0028&32.75&0.9697$\pm$0.0031&43.20\\ 
$data_8$&0.9901$\pm$0.0013&9.96&0.9290$\pm$0.0045&119.64&0.9263$\pm$0.0069&104.94\\
$data_9$&0.9899$\pm$0.0028&1.02&0.9249$\pm$0.0096&68.91&0.9264$\pm$0.0067&123.85\\
\end{tabular}
\end{small}
\end{center}
\vskip -0.1in
\end{table}

\begin{table}[H]    
\centering
 \caption{$F_{error}(w)$ vs. $F_{log}(w)$ minimization via Algorithm \ref{GD} on real data sets.}
   \vskip 0.1in
  \label{t_0-1_2}
   \vskip 0.05in
\begin{center}
\begin{small}
 \begin{tabular}{c|cc|cc}  
\multirow{2}{*}{Data} 
&\multicolumn{2}{c|}{\textbf{$\pmb{F_{error}(w)}$ Minimization}}&\multicolumn{2}{c}{\textbf{$\pmb{F_{log}(w)}$ Minimization}}\\   
&{Accuracy$\pm$ std}&{Time (s)}&{Accuracy $\pm$ std}&{Time (s)}\\
\hline
fourclass&0.8782$\pm$0.0162&0.02&0.8800$\pm$0.0147&0.12\\ 
svmguide1&\textbf{0.9735}$\pm$0.0047&0.42&0.9506$\pm$0.0070&0.28\\ 
diabetes&0.8832$\pm$0.0186&1.04&0.8839$\pm$0.0193&0.13\\ 
shuttle&0.8920$\pm$0.0015&0.01&\textbf{0.9301}$\pm$0.0019&4.05\\ 
vowel&0.9809$\pm$0.0112&0.91&0.9826$\pm$0.0088&0.11\\ 
magic04&0.8867$\pm$0.0044&0.66&0.8925$\pm$0.0041&1.75\\  
poker&0.9897$\pm$0.0008&0.17&0.9897$\pm$0.0008&10.96\\ 
letter&0.9816$\pm$0.0015&0.01&0.9894$\pm$0.0009&4.51\\ 
segment&0.9316$\pm$0.0212&0.17&\textbf{0.9915}$\pm$0.0101&0.36\\ 
svmguide3&\textbf{0.9118}$\pm$0.0136&0.39&0.8951$\pm$0.0102&0.17\\ 
ijcnn1&0.9512$\pm$0.0011&0.01&0.9518$\pm$0.0011&4.90\\  
german&0.8780$\pm$0.0125&1.09&0.8826$\pm$0.0159&0.62\\ 
landsat satellite&0.9532$\pm$0.0032&0.01&0.9501$\pm$0.0049&3.30\\ 
sonar&\textbf{0.8926}$\pm$0.0292&0.49&0.8774$\pm$0.0380&0.92\\ 
a9a&0.9193$\pm$0.0021&0.98&0.9233$\pm$0.0020&11.45\\ 
w8a&0.9851$\pm$0.0005&0.36&0.9876$\pm$0.004&24.16\\ 
mnist&0.9909$\pm$0.0004&3.79&0.9938$\pm$0.0004&136.83\\ 
colon cancer &\textbf{0.9364}$\pm$0.0394&15.92&0.8646$\pm$0.0555&1.20\\ 
gisette&0.9782$\pm$0.0025&310.72&0.9706$\pm$0.0036&136.71\\
\end{tabular}
\end{small}
\end{center}
\vskip -0.1in
\end{table}

\begin{table}[H]     
\centering 
 \caption{$F_{AUC}(w)$ vs. $F_{hinge}(w)$ minimization via Algorithm \ref{GD} on artificial data sets.}
   \vskip 0.1in
  \label{t_art_AUC_2}
   \vskip 0.05in
\begin{center}
\begin{small}
 \begin{tabular}{ c|cc|cc|cc}  
\multirow{3}{*}{Data} 
&\multicolumn{2}{c|}{\textbf{$\pmb{F_{AUC}(w)}$ Minimization}}&\multicolumn{2}{c|}{\textbf{$\pmb{F_{AUC}(w)}$ Minimization}}&\multicolumn{2}{c}{\textbf{$\pmb{F_{hinge}(w)}$ Minimization}}\\  
&\multicolumn{2}{c|}{\textbf{Exact moments}}&\multicolumn{2}{c|}{\textbf{Approximate moments}}&\\ 
&{AUC$\pm$ std}&{Time (s)}&{AUC $\pm$ std}&{Time (s)}&{AUC $\pm$ std}&{Time (s)}\\  
\hline
$data_1$&0.9972$\pm$0.0014&0.01&\textbf{0.9941}$\pm$0.0027&0.23&0.9790$\pm$0.0089&5.39\\
$data_2$&0.9963$\pm$0.0016&0.01&\textbf{0.9956}$\pm$0.0018&0.22&0.9634$\pm$0.0056&159.23\\
$data_3$&0.9965$\pm$0.0015&0.01&\textbf{0.9959}$\pm$0.0018&0.24&0.9766$\pm$0.0041&317.44\\
$data_4$&0.9957$\pm$0.0018&0.02&\textbf{0.9933}$\pm$0.0022&0.83&0.9782$\pm$0.0054&23.36\\
$data_5$&0.9962$\pm$0.0011&0.02&\textbf{0.9951}$\pm$0.0013&0.80&$0.9589\pm$0.0068&110.26\\
$data_6$&0.9962$\pm$0.0013&0.02&\textbf{0.9949}$\pm$0.0015&0.82&0.9470$\pm$0.0086&275.06\\
$data_7$&0.9965$\pm$0.0021&0.08&\textbf{0.9874}$\pm$0.0034&4.61&0.9587$\pm$0.0092&28.31\\ 
$data_8$&0.9966$\pm$0.0008&0.07&\textbf{0.9929}$\pm$0.0017&4.54&0.9514$\pm$0.0051&104.16\\
$data_9$&0.9962$\pm$0.0014&0.08&\textbf{0.9932}$\pm$0.0020&4.54&0.9463$\pm$0.0085&157.62\\
\end{tabular}
\end{small}
\end{center}
\vskip -0.1in
\end{table}

\begin{table}[H]      
\centering
\caption{$F_{AUC}(w)$ vs. $F_{hinge}(w)$ minimization via Algorithm \ref{GD} on real data sets.}
   \vskip 0.1in
   \label{t_AUC_2}
   \vskip 0.05in
\begin{center}
\begin{small}
 \begin{tabular}{c|cc|cc}  
\multirow{2}{*}{Data} 
&\multicolumn{2}{c|}{\textbf{$\pmb{F_{AUC}(w)}$ Minimization}}&\multicolumn{2}{c}{\textbf{$\pmb{F_{hinge}(w)}$ Minimization}}\\   
&{AUC$\pm$ std}&{Time (s)}&{AUC $\pm$ std}&{Time (s)}\\  
\hline
fourclass&0.8362$\pm$0.0312&0.01&0.8362$\pm$0.0311&6.81\\ 
svmguide1&0.9717$\pm$0.0065&0.06&0.9863$\pm$0.0037&35.09\\ 
diabetes&0.8311$\pm$0.0311&0.03&0.8308$\pm$0.0327&12.48\\ 
shuttle&0.9872$\pm$0.0013&0.11&0.9861$\pm$0.0017&2907.84\\ 
vowel&0.9585$\pm$0.0333&0.12&\textbf{0.9765}$\pm$0.0208&2.64\\ 
magic04&0.8382$\pm$0.0071&0.11&0.8419$\pm$0.0071&1391.29\\  
poker&0.5054$\pm$0.0224&0.11&0.5069$\pm$0.0223&1104.56\\ 
letter&0.9830$\pm$0.0029&0.12&0.9883$\pm$0.0023&121.49\\ 
segment&0.9948$\pm$0.0035&0.21&0.9992$\pm$0.0012&4.23\\ 
svmguide3&\textbf{0.8013}$\pm$0.0420&0.34&0.7877$\pm$0.0432&23.89\\ 
ijcnn1&0.9269$\pm$0.0036&0.08&0.9287$\pm$0.0037&2675.67\\  
german&0.7938$\pm$0.0292&0.14&0.7919$\pm$0.0294&32.63\\ 
landsat satellite&\textbf{0.7587}$\pm$0.0160&0.43&0.7458$\pm$0.0159&193.46\\ 
sonar&0.8214$\pm$0.0729&0.88&\textbf{0.8456}$\pm$0.0567&2.15\\ 
a9a&0.9004$\pm$0.0039&0.92&0.9027$\pm$0.0037&15667.87\\ 
w8a&0.9636$\pm$0.0055&0.54&0.9643$\pm$0.0057&5353.23\\ 
mnist&0.9943$\pm$0.0009&0.64&0.9933$\pm$0.0009&28410.2393\\ 
colon cancer&\textbf{0.8942}$\pm$0.1242&2.50&0.8796$\pm$0.1055&0.05\\ 
gisette&\textbf{0.9957}$\pm$0.0015&31.32&0.9858$\pm$0.0029&3280.38\\ 
\end{tabular}
\end{small}
\end{center}
\vskip -0.1in
\end{table}

\section{Numerical Comparison vs. LDA}
In the following we provide the numerical results comparing minimizing $F_{error}(w)$ and $F_{log}(w)$ versus LDA, while using the artificial data sets as well as real data sets.

\begin{table}[H]   
\centering
 \caption{$F_{error}(w)$ and $F_{log}(w)$ minimization via Algorithm \ref{GD} vs. LDA on artificial data sets.}
   \vskip 0.1in
  \label{app_t_1}
   \vskip 0.05in
\begin{center}
\begin{small}
 \begin{tabular}{ c|c|c|c|c}  
\multirow{3}{*}{Data} 
&{\textbf{$\pmb{F_{error}(w)}$ Minimization}}&{\textbf{$\pmb{F_{error}(w)}$ Minimization}}&{\textbf{$\pmb{F_{log}(w)}$ Minimization}}&{\textbf{LDA}}\\  
&{\textbf{Exact moments}}&{\textbf{Approximate moments}}&&\\ 
&{Accuracy$\pm$ std}&{Accuracy $\pm$ std}&{Accuracy $\pm$ std}&{Accuracy $\pm$ std}\\
\hline
$data_1$&0.9965$\pm$0.0008&0.9907$\pm$0.0014&0.9897$\pm$0.0018&0.9851$\pm$0.0035\\ 
$data_2$&0.9905$\pm$0.0023&\textbf{0.9806}$\pm$0.0032&0.9557$\pm$0.0049&0.9670$\pm$0.0057\\
$data_3$&0.9884$\pm$0.0030&\textbf{0.9745}$\pm$0.0037&0.9537$\pm$0.0048&0.9630$\pm$0.0081\\ 
$data_4$&0.9935$\pm$0.0017&0.9791$\pm$0.0034&0.9782$\pm$0.0031&0.9672$\pm$0.0049\\
$data_5$&0.9899$\pm$0.0026&\textbf{0.9716}$\pm$0.0048&0.9424$\pm$0.0055&0.9455$\pm$0.0074\\
$data_6$&0.9904$\pm$0.0017&\textbf{0.9670}$\pm$0.0058&0.9291$\pm$0.0076&0.9417$\pm$0.0085\\
$data_7$&0.9945$\pm$0.0019&0.9786$\pm$0.0028&0.9697$\pm$0.0031&0.9086$\pm$0.0137\\ 
$data_8$&0.9901$\pm$0.0013&0.9290$\pm$0.0045&0.9263$\pm$0.0069&0.8526$\pm$0.0182\\
$data_9$&0.9899$\pm$0.0028&0.9249$\pm$0.0096&0.9264$\pm$0.0067&0.8371$\pm$0.0149\\
\end{tabular}
\end{small}
\end{center}
\vskip -0.1in
\end{table}

\newpage
\begin{table}[H]    
\centering
 \caption{$F_{error}(w)$ and $F_{log}(w)$ minimization via Algorithm \ref{GD} vs. LDA  on real data sets.}
   \vskip 0.1in
  \label{app_t_2}
   \vskip 0.05in
\begin{center}
\begin{small}
 \begin{tabular}{c|cc|cc|cc}  
\multirow{2}{*}{Data} 
&{\textbf{$\pmb{F_{error}(w)}$ Minimization}}&{\textbf{$\pmb{F_{log}(w)}$ Minimization}}&{\textbf{LDA}}\\   
&{Accuracy$\pm$ std}&{Accuracy $\pm$ std}&{Accuracy $\pm$ std}\\
\hline
fourclass&0.8782$\pm$0.0162&0.8800$\pm$0.0147&0.7572$\pm$0.0314\\ 
svmguide1&\textbf{0.9735}$\pm$0.0047&0.9506$\pm$0.0070&0.8972$\pm$0.0159\\ 
diabetes&0.8832$\pm$0.0186&0.8839$\pm$0.0193&0.7703$\pm$0.0366\\ 
shuttle&0.8920$\pm$0.0015&\textbf{0.9301}$\pm$0.0019&0.9109$\pm$0.0027\\ 
vowel&0.9809$\pm$0.0112&0.9826$\pm$0.0088&0.9600$\pm$0.0224\\ 
magic04&0.8867$\pm$0.0044&0.8925$\pm$0.0041&0.7841$\pm$0.0093\\  
poker&0.9897$\pm$0.0008&0.9897$\pm$0.0008&0.9795$\pm$0.0017\\ 
letter&0.9816$\pm$0.0015&0.9894$\pm$0.0009&0.9711$\pm$0.0029\\ 
segment&0.9316$\pm$0.0212&\textbf{0.9915}$\pm$0.0101&0.9617$\pm$0.0331\\ 
svmguide3&\textbf{0.9118}$\pm$0.0136&0.8951$\pm$0.0102&0.8238$\pm$0.0259\\ 
ijcnn1&0.9512$\pm$0.0011&0.9518$\pm$0.0011&0.9081$\pm$0.0029\\  
german&0.8780$\pm$0.0125&0.8826$\pm$0.0159&0.7675$\pm$0.0275\\ 
landsat satellite&0.9532$\pm$0.0032&0.9501$\pm$0.0049&0.9061$\pm$0.0065\\ 
sonar&\textbf{0.8926}$\pm$0.0292&0.8774$\pm$0.0380&0.7622$\pm$0.0499\\ 
a9a&0.9193$\pm$0.0021&0.9233$\pm$0.0020&0.8452$\pm$0.0038\\ 
w8a&0.9851$\pm$0.0005&0.9876$\pm$0.004&0.9839$\pm$0.0012\\ 
mnist&0.9909$\pm$0.0004&0.9938$\pm$0.0004&0.9778$\pm$0.0013\\ 
colon cancer &\textbf{0.9364}$\pm$0.0394&0.8646$\pm$0.0555&0.8875$\pm$0.0985\\ 
gisette&0.9782$\pm$0.0025&0.9706$\pm$0.0036&0.5875$\pm$0.0207\\
\end{tabular}
\end{small}
\end{center}
\vskip -0.1in
\end{table}

\end{document}